\documentclass[10pt,twocolumn]{article} 
\usepackage{simpleConference}
\usepackage{times}
\newtheorem{definition}{Definition}
\usepackage{graphicx}
\usepackage{amssymb}
\usepackage{url,hyperref}
\usepackage{caption}
\usepackage{subcaption}
\usepackage{amsmath,amssymb,amsfonts}
\usepackage{algorithm}
\usepackage{multirow}
\usepackage{bbding}
\usepackage{algpseudocode}
\newcommand*\samethanks[1][\value{footnote}]{\footnotemark[#1]}

\begin{document}

\title{\Large Subgraph Centralization:\\ A Necessary Step for Graph Anomaly Detection}

\author{Zhong Zhuang\thanks{National Key Laboratory for Novel Software Technology, Nanjing University, \{zhuangz,tingkm,songsb\}@lamda.nju.edu.cn} \and Kai Ming Ting\samethanks[1] \and Guansong Pang\thanks{Singapore Management University, gspang@smu.edu.sg} \and Shuaibin Song\samethanks[1]}

\maketitle
\thispagestyle{empty}

\begin{abstract}
Graph anomaly detection has attracted a lot of interest recently. Despite their successes, existing detectors have at least two of the three weaknesses: (a) high computational cost which limits them to small-scale networks only; (b) existing treatment of subgraphs produces suboptimal detection accuracy; and (c) unable to provide an explanation as to why a node is anomalous, once it is identified. We identify that the root cause of these weaknesses is a lack of a proper treatment for subgraphs. A treatment called Subgraph Centralization for graph anomaly detection is proposed to address all the above weaknesses. Its importance is shown in two ways. First, we present a simple yet effective new framework called Graph-Centric Anomaly Detection (GCAD). The key advantages of GCAD over existing detectors including deep-learning detectors are: (i) better anomaly detection accuracy; (ii) linear time complexity with respect to the number of nodes; and (iii) it is a generic framework that admits an existing point anomaly detector to be used to detect node anomalies in a network. Second, we show that Subgraph Centralization can be incorporated into two existing detectors  to overcome the above-mentioned weaknesses.
\end{abstract}

\section{Introduction}
\label{sec:introduction}

\begin{figure}[t]

 \center
   
  \includegraphics[width=\columnwidth]{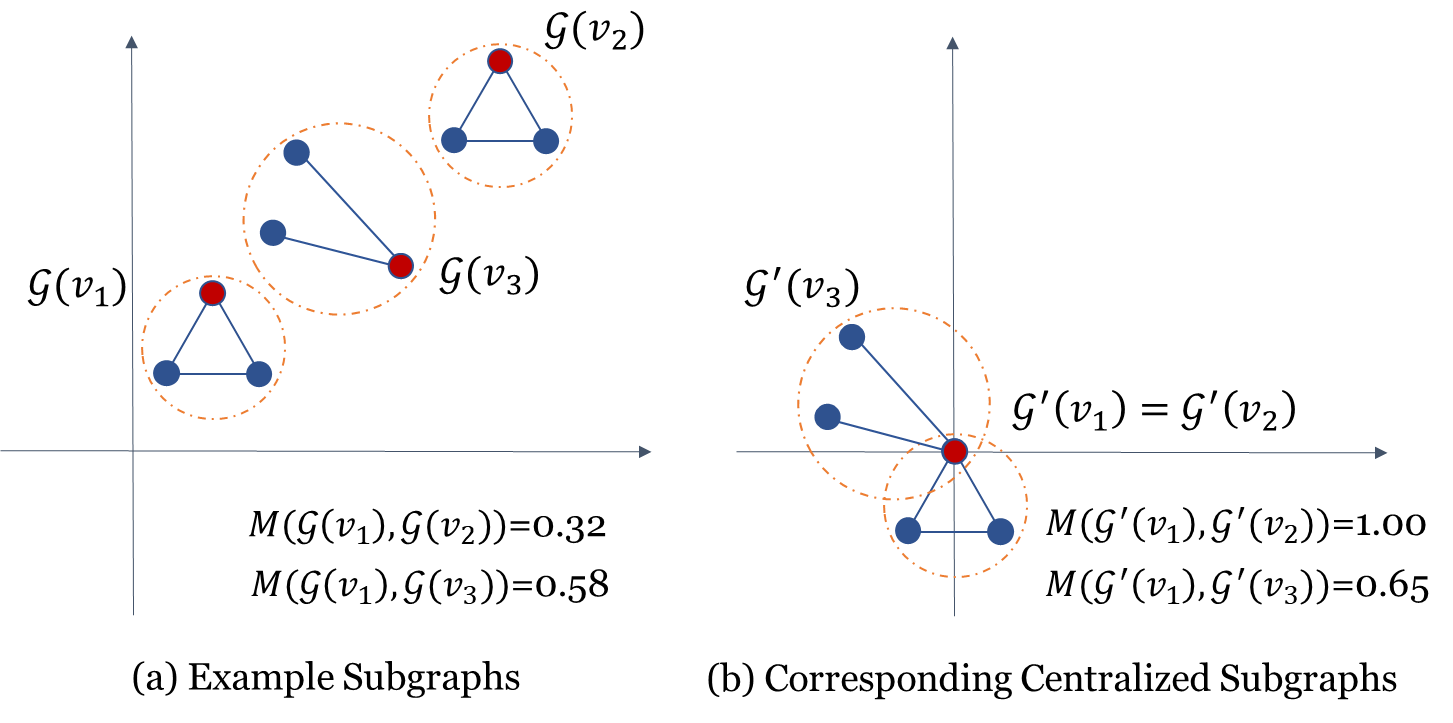}
  \caption{(a) Three example subgraphs of a (not-shown) network in the space of given node vectors. The red node represents the source node $v$ of a subgraph $\mathcal{G}(v)$. $M$ denotes a similarity measure between subgraphs. (b) In the translated space, each centralized subgraph $\mathcal{G}'(v)$ has all its nodes translated to the same extent as source node $v$ is translated to the origin. $\mathcal{G}'(v_1)=\mathcal{G}'(v_2)$ because they have the same structure.}
   \label{fig:centralization}
 
\end{figure}

Attributed networks are omnipresent in various applications because they are a powerful means to represent node information as well as the relationship between nodes well. Examples are: (a) In a social network \cite{bird} (e.g., Facebook, Blog, LinkedIn), users with attribute information following  or being followed by other users can be seen as nodes and connections in an attributed network. (b) In a citation network \cite{10.1145/3159652.3159655} of scientific publications, authors denote nodes, and the connections between nodes denote joint publications between authors. 

Anomalous node detection in an attributed network has many applications, e.g., fraud detection, social spam detection and academic misconduct detection. 

Existing anomalous node detectors can be categorized into two approaches, i.e., full-graph-based detectors and subgraph-based detectors. The former has high computational cost which limits them to small-scale networks only. We discover that the latter does not use subgraphs to their full potential resulting in suboptimal detection accuracy. In addition, both types of detectors are unable to provide an explanation as to why a node is anomalous, once it is identified.

The importance of subgraphs can be understood from the fact that each node $v$ is influenced by all the nodes connected to it directly and transitively in a subgraph centered at $v$. The nodes connected via many hops outside the subgraph have negligible or no influence. Therefore, the similarity between subgraphs determine whether a subgraph is normal or anomalous.

Though the second existing approach has the same understanding, the methodology has  a serious weakness, i.e., no centralization is performed before computing the similarity between subgraphs. As a result, the similarity between subgraphs can be misleading. An example is shown in Figure \ref{fig:centralization}(a): subgraph $\mathcal{G}(v_1)$ is deemed to be more similar to $\mathcal{G}(v_3)$ than $\mathcal{G}(v_2)$ simply because the node positions of the first two graphs are closer, i.e., the node positions play a more important role than the graph structures in the similarity calculation.

Intuitively, the similarity between subgraphs rely primarily on the graph structure, and the absolute positions in the space of the given node vectors shall play no role in the calculation. This can be achieved by translating all nodes of a subgraph $\mathcal{G}(v)$ to the same extent as the source node $v$ is translated to the origin. Figure~\ref{fig:centralization}(b) shows the translated subgraphs (in the translated space) of those shown in Figure \ref{fig:centralization}(a) (in the original space). Indeed, the similarity between the translated subgraphs $\mathcal{G}'(v_1)$ and $\mathcal{G}'(v_2)$ is unity because they have the same structure; and each of the them is less similar to $\mathcal{G}'(v_3)$ due to the difference in structures.

We call this technique {\em Subgraph Centralization}. We show that it is a necessary step in a proposed new graph anomaly detection framework and in  addressing all the above-mentioned weaknesses of existing detectors.

Our main contributions are summarized as follows:
\renewcommand{\theenumi}{\arabic{enumi}}
\begin{enumerate}
    \item Formally define normal and anomalous nodes in a network. As far as we know, we are the first to provide such definitions on graph anomaly detection.
    \item Propose a simple but effective new framework called Graph-Centric Anomaly Detection or GCAD which employs Subgraph Centralization as a necessary step. GCAD is capable of detecting anomaly on large-scale networks and providing an explanation why a node is anomalous/normal. The framework also admits an existing point anomaly detector to be used to perform graph anomaly detection.
    \item Demonstrate that Subgraph Centralization can be applied to two existing detectors to improve their detection accuracy, and empower their ability to explain their predictions.
\end{enumerate}

GCAD is distinguished from the two existing approaches with two unique features, i.e., the use of centralized subgraphs as the basis for similarity measurement and explainability.  Subgraphs are the key to the definition of node anomalies and the design of the GCAD framework. The centralization is simple yet crucial to the success of GCAD. The explainability of GCAD depends on the centralized subgraphs, but independent of the point anomaly detector used in GCAD.

\section{Related Work}
\label{sec:related}
To detect anomalies, it is vital to model some characteristics of a network that represent the normality of the network. 
We categorize existing works into two approaches, along the line whether the entire network or subgraphs are used in the modeling:\\
The full-graph-based approach is to model some form of normal characteristics from an entire network. ANOMALOUS \cite{ijcai2018-488} selects attributes on the space of features based on the structure of a network and applies residual analysis to detect anomalies. DOMINANT \cite{ding2019deep} combines GCN \cite{Kipf:2016tc} with an autoencoder to reconstruct the network (via both the attribute matrix and adjacency matrix) such that the normal nodes are  reconstructed with small errors. A few other works \cite{9053387, 10.1145/3357384.3358074} have followed the same methodology with some improvements. Oddball \cite{10.1007/978-3-642-13672-6_40} models the normal characteristics of a network via a power-law model. OCGNN \cite{OCGNN} applies GCN and hypersphere learning \cite{SVDD} to perform representation learning on a network.\\
The subgraph-based approach utilizes various ways to extract subgraphs from a network. CoLA \cite{liu2021anomaly} generates a large set of positive and negative subgraphs, where a normal subgraph depicts the normal relationship between each node and its neighbouring structure in the network; and a negative subgraph does not. This set is used to train a classifier. ANEMONE \cite{jin2021anomaly} and SL-GAD \cite{SL-GAD} employ self-supervised learning by generating subgraphs surrounding target nodes and performing patch-level and contextual-level contrastive learning.\\
However, none of them utilize Subgraph Centralization, leading to the problem we mentioned in Section \ref{sec:introduction}.

\section{Node Anomaly: Definitions}
\label{sec:definition}
Let $\mathcal{G} = \{\mathcal{V}, \mathcal{E}, \mathcal{X}\}$ be an undirected attributed network, where $\mathcal{V} = \{v_1, \dots, v_n\}$ denotes the set of nodes, $\mathcal{E}$ denotes the set of edges, $\mathcal{X} \in \mathbb{R}^{n \times d}$ denotes the attribute matrix of nodes,  and each node vector has $d$ attributes. 
 
 Node anomaly detection is defined as:
\begin{definition}[Node Anomaly Detection]
The task of node anomaly detection in a network $\mathcal{G} = \{\mathcal{V}, \mathcal{E}, \mathcal{X}\}$ is to identify the few node anomalies which have characteristics different from the majority of the nodes in the network, and to rank all the nodes on the basis of their anomaly scores such that the node anomalies are ranked higher than the normal ones.
\end{definition}

To operationalize the above generic definition, a method must be devised to provide a score for each node. We propose to use $h$-subgraphs as the basic means to do this in the next subsection.

\subsection{Node Anomalies Based on $h$-subgraphs}
\label{definition}

\begin{definition}[$h$-subgraph]
An $h$-subgraph $\mathcal{G}^{h}(v) = (\mathbf{V}, \mathbf{E}, \mathbf{X})$ is a subgraph rooted at source node $v$ such that the shortest path between any node in $\mathcal{G}^{h}(v)$  and $v$ has length $\le h$, where $h  \in \mathbb{N}$ is the maximum depth of the subgraph.
\label{def:subgraph}
\end{definition}

Given a network $\mathcal{G} = \{\mathcal{V}, \mathcal{E}, \mathcal{X}\}$ having $n$ nodes, there are a total of $n$ $h$-subgraphs. Among the $n$ $h$-subgraphs, the ones which are few and different in terms of some measure from the majority of the $h$-subgraphs are regarded as anomalous; and the source nodes of these $h$-subgraphs are regarded as node anomalies in $\mathcal{G}$.
We propose to use a similarity measure to characterize the differences between $h$-subgraphs.

The node anomaly based on $h$-subgraphs is defined as follows.
Let the set of nodes $\mathcal{V}$ consists of two subsets $\mathcal{V}^{A}$ and $\mathcal{V}^{N}$ which denote the sets of anomalous nodes and normal nodes, respectively, where $|\mathcal{V}^{N}| \gg |\mathcal{V}^{A}|$, and $M(\cdot,\cdot)$ be a similarity measure between two $h$-subgraphs.

\begin{definition}Node $u$ in $\mathcal{G} = \{\mathcal{V}, \mathcal{E}, \mathcal{X}\}$ is
\begin{itemize}
    \item  A normal node if $\mathcal{G}^{h}(u)$ is similar to most other $h$-subgraphs, i.e., $\frac{1}{|\mathcal{V}|} \sum_{v \in \mathcal{V}} M(\mathcal{G}^{h}(u), \mathcal{G}^{h}(v))$ is large.
    \item An  anomalous node if $\mathcal{G}^{h}(u)$ is dissimilar to most other $h$-subgraphs, i.e., $\frac{1}{|\mathcal{V}|} \sum_{v \in \mathcal{V}} M(\mathcal{G}^{h}(u), \mathcal{G}^{h}(v))$ is small.
\end{itemize}
\label{def:normal}
\end{definition}
In practice, the difference can be parameterized as:
$u$ is an anomalous node if $\frac{1}{|\mathcal{V}|} \sum_{v \in \mathcal{V}} M(\mathcal{G}^{h}(u), \mathcal{G}^{h}(v)) < \tau$; otherwise $u$ is a normal node. Or simply select the top  \texttt{m} anomalous nodes which have the lowest similarities.

\section{Proposed Framework: GCAD}
\label{sec:GCAD}

We propose a new framework called GCAD (Graph-Centric node Anomaly Detection) to detect node anomalies in a network. GCAD scores every node  by examining the similarity between  $h$-subgraphs extracted from a network. A node $v$ has a high score if the $h$-subgraph with $v$ as the source node  is dissimilar to most other $h$-subgraphs in the network. 

The proposed framework consists of four main components: subgraph extraction and centralization, subgraph embedding, anomaly detection and subgraph depth-based weighted anomaly scoring, as shown in Algorithm \ref{alg:framework}.

\begin{algorithm}
\caption{GCAD}
\label{alg:framework}
\begin{algorithmic}[1]
\Require Network: $\mathcal{G}=(\mathcal{V}, \mathcal{E}, \mathcal{X})$;
subgraph depth: $h$; the parameter of depth-based weighted score: $\lambda$
\Ensure Anomaly Scores $\hat{Y}$ for all $v\in \mathcal{V}$
\State $\mathcal{S}\leftarrow$SEC($\mathcal{G}$, $h$);  \small{(Extract \& Centralize Subgraphs)}
\For{each $\mathcal{G}^{h}(v)\in \mathcal{S}$}
\State $\mathbf{e}_{v} = \phi(\mathcal{G}^{h}({v}))$;  \small{(Embed each $\mathcal{G}^{h}$ to a vector)}
\EndFor
\State Let $\mathbf{E}$ be the  matrix of embedded vectors $\mathbf{e}_{v}$ for all $v \in \mathcal{V}$
\State $Y \leftarrow \text{Detector}(\mathbf{E})$; \small{(Score $v \in \mathcal{V}$ with a point detector)}
\State Compute the reweighted scores $\hat{Y}$ via Eq. \ref{eq:1};

\State \Return $\hat{Y}$
\end{algorithmic}
\end{algorithm}

As the base unit of operation is $h$-subgraphs, the algorithm begins to extract $n$ $h$-subgraphs from a given network consisting of $n$ nodes. 

To enable similarity measurements among $h$-subgraphs, one key step is to centralize all these $h$-subgraphs to nullify the unwanted interference of node positions in the node vector space, since we are not interested in their absolute difference in the node vector space. Only then it is meaningful to examine the (dis)similarity between $h$-subgraphs, purely based on the structure of individual $h$-subgraphs.  
This process is subgraph centralization. Both the extraction and centralization are conducted in line\#1 in Algorithm \ref{alg:framework}). 

Then, an embedding method is required to generate an embedded representation for each $h$-subgraph (line\#3 in Algorithm \ref{alg:framework}). An existing point anomaly detector can then be applied on the set of embedded vectors to detect node anomalies which are different from the majority in the set, as shown in line\#6 in Algorithm \ref{alg:framework}. Since each $h$-subgraph is derived from a source node, the anomaly score of each $h$-subgraph denotes the anomaly score of its source node. 

Finally, to derive the final score of a node $u$ (line\#6 in Algorithm~\ref{alg:framework}), a weighted anomaly scoring method is used to aggregate the anomaly scores of source nodes of all $h$-subgraphs which contain node $u$.

We provide the details of the four steps in the following four subsections.

\begin{figure*}

 \center

  \includegraphics[width=\textwidth]{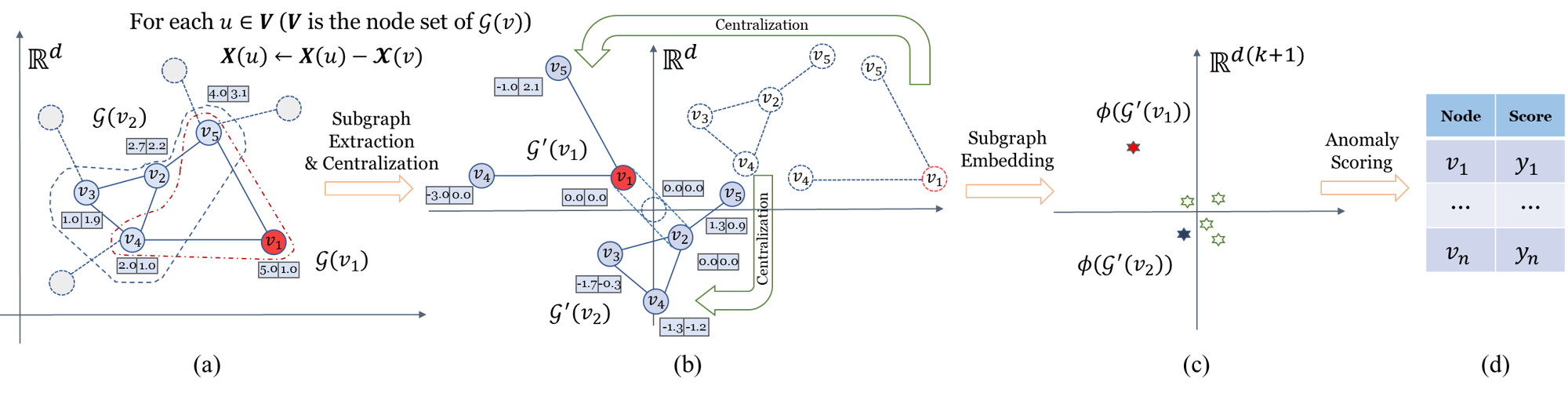}

  \caption{
  Overview of the GCAD framework. Only the first three components are shown here, i.e., $h$-subgraph extraction and centralization, subgraph embedding, and anomaly scoring. (a) The red node $v_1$ in the node vector space $\mathbb{R}^{d}$ is an anomalous node because it has connections with two distant nodes $v_4$ and $v_5$ while the other nodes are connected to neighbouring nodes only. Each dotted loop indicates an $h$-subgraph, extracted from the given network, where $h=1$. (b) The two example $1$-subgraphs are centralized such that their source nodes are translated to the origin. For clarity, the source nodes $v_1$ and $v_2$ are drawn besides the origin without overlapping (two parallel doted lines extended from the origin). Subfigure (c) shows the embeddings of the centralized subgraphs in the embedded $\mathbb{R}^{d(k+1)}$ space. $\phi(\mathcal{G}(v_1))$ is an outlier which is different from other points in this space. 
  }

  \label{Framework}

\end{figure*}

\subsection{Subgraph Extraction and Centralization (SEC).}
Figure \ref{Framework}a provides an illustration of the subgraph extraction process. Given an $h$ setting, extracting $h$-subgraphs from a given network is straightforward: each node in the network is treated as a source node $u$; and the $h$-subgraph with source node $u$ is extracted via connected nodes up to depth $h$. The two 1-subgraphs are in the top-right corner in Figure \ref{Framework}b.

The aim, according to Definition \ref{def:normal}, is to compare the structures of individual $h$-subgraphs which are determined from the relative positions of nodes in an $h$-subgraph, but independent of the absolute positions of the nodes in the node vector space. To achieve this aim, the subgraph centralization component  maps all source nodes of $h$-subgraphs to the origin of the node vector space. To ensure that the structure of every $h$-subgraph remains unchanged,  every node in an $h$-subgraph is translated to the same extent in which its source node is translated to the origin. This centralization nullifies the unwanted interference of the original positions in the node vector space, and enables only the structures of the $h$-subgraphs to be compared. The process is illustrated in Figure~\ref{Framework}b. The detailed procedure of the above two processes is shown in Algorithm \ref{alg:Subgraph}.

\begin{algorithm}
\caption{SEC}
\label{alg:Subgraph}

\begin{algorithmic}[1]
\Require Network: $\mathcal{G}=(\mathcal{V}, \mathcal{E}, \mathcal{X})$; subgraph depth: $h$
\Ensure Set of centralized $h$-subgraphs $\mathcal{S}$
\State Initialize $\mathcal{S}=\{\}$;
\For{each $v \in \mathcal{V}$}
\State Extract $\mathcal{G}^{h}(v) = (\mathbf{V}, \mathbf{E}, \mathbf{X})$;\small{(subgraph extraction)}
\For{each $u \in \mathbf{V}$} \small{($\mathbf{V}$ includes the source node $v$)}
\State $\mathbf{X}(u) \leftarrow \mathbf{X}(u) - \mathcal{X}(v)$; \small{(Centralization)}
\EndFor
\State $\mathcal{S} \leftarrow \mathcal{S} \cup \{\mathcal{G}^{h}(v)\}$;
\EndFor
\State \Return $\mathcal{S}$
\end{algorithmic}
\end{algorithm}

Once the above process is completed, the majority of centralized $h$-subgraphs which are similar to each other constitute the normal $h$-subgraphs in a network. The few centralized $h$-subgraphs which are dissimilar to the normal ones are anomalies.

To facilitate such a comparison, we need to embed each $h$-subgraph into a vector, which is the topic of the next subsection. For brevity, we use $\mathcal{G}^{h}(v)$ to denote a centralized $h$-subgraph with source node $v$ hereafter.

\subsection{Subgraph Embedding.}
A subgraph embedding method $\phi$ to map each centralized $h$-subgraph $\mathcal{G}^{h}(v)$ into a vector $\phi(\mathcal{G}^{h}(v))$. We use the widely-used subgraph embedding method, Weisfeiler-Lehman (WL) \cite{Togninalli19,Xu_Ting_Jiang_2021}. More specifically, 
for every $h$-subgraph $\mathcal{G}^{h}(v)=\{\mathbf{V}, \mathbf{E}, \mathbf{X}\}$, we utilize the WL scheme in the node vector space to get the embedding for every node of an $h$-subgraph. The key idea is to create an explicit propagation scheme that leverages and iteratively updates the current node vector by averaging over the node vectors in the neighbourhoods. 

Let the node vector $\mathbf{X}^{0}(u) = \mathbf{X}(u) \in \mathbb{R}^d$ for each node $u \in \mathbf{V}$. The node vector of the $k$th iteration $\mathbf{X}^{k}(u)$, for $k \ge 1$, is computed via WL as follows:
\begin{equation}
    \mathbf{X}^{k}(u) = \frac{1}{2}\left(\mathbf{X}^{k - 1}(u) + \frac{1}{deg(u)}\sum_{w\in \mathbf{N}(u)}\mathbf{X}^{k - 1}(w)\right),
\end{equation}
where $deg(u)$ denotes the degree of node $u$, and $\mathbf{N}(u)$ is set of the one-hop neighbours of node $u$ in the $h$-subgraph $\mathcal{G}^{h}(v)$.

\begin{definition} The WL-embedded vector of a node $u$ in an $h$-subgraph
    $\mathcal{G}^{h}(v)=\{\mathbf{V}, \mathbf{E}, \mathbf{X}\}$ is defined as $\phi(u) \in \mathbb{R}^{d(k+1)}$:
    \begin{equation}
        \phi(u) = [\mathbf{X}^{0}(u), \dots, \mathbf{X}^{k}(u)]^{\top}.
    \end{equation}
\end{definition}
Note that the concatenation of embedding from each iteration is different from other GNN-based methods in that GNN-based embedding does not concatenate outputs of different layers.

\begin{definition}
The WL-embedded vector $\mathbf{e}_v$ of the entire $h$-subgraph $\mathcal{G}^{h}(v)$ is defined as the mean embedding of all nodes in the $h$-subgraph:
    \begin{equation}
        \mathbf{e}_v = \phi(\mathcal{G}^{h}(v)) = \frac{1}{|\mathbf{V}|}\sum_{u \in \mathbf{V}}\phi(u).
    \end{equation}
    
\end{definition}

Because the largest $k$ for an $h$-subgraph is $h$. Therefore, we have set $k=h$.

\subsection{Point Anomaly Detector.}

Given a set of embedded vectors, an existing unsupervised anomaly detector can then be used to detect anomalies that are different from the majority in the set.

Let $\mathbf{E}$ be the matrix of the embedded vectors $\mathbf{e}_v$. We assume that $Detector(\mathbf{E})$ is trained from $\mathbf{E}$ and produces output $Y$ which is a one-dimensional matrix of anomaly scores $y_v$ for all nodes $v$ in the network. Any existing point anomaly detector can be used here.

\subsection{Depth-based Weighted Score.}
The point anomaly detector in the last subsection produces a score $y_v$ for node $v$ which is the source node of $h$-subgraph $\mathcal{G}^{h}(v)$ for every node in a given network.
The score can be used directly as the anomaly score of each node. But, we introduce a depth-based weighted score here for further improvement. 

Let $u$ be the node of interest; and $\mathcal{V}_u$ be the set of all source nodes of $h$-subgraphs which contain $u$. The depth-based weighted score aggregates scores from all $h$-subgraphs in $\mathcal{V}_u$. The idea is to place a larger weight (less than 1) to the score $y_v$ of $\mathcal{G}^{h}(v)$ if $u$ is closer to $v$. 

The weight is formulated using a parameter $\lambda$ as $\lambda^{\ell(v, u)}$,
where $0 \leq \lambda < 1$; and $\ell(v,u)$ is the number of hops from $v$ to $u$ in $\mathcal{G}^{h}(v)$.

The final anomaly score $\hat{y}_u$ for node $u$ is defined as:
\begin{equation}\label{eq:1}
    \hat{y}_u = \frac{\sum_{v \in \mathcal{V}_u} \lambda^{\ell(v, u)} y_v  }{\sum_{v \in \mathcal{V}_u} \lambda^{\ell(v, u)}}.
\end{equation}

\section{Applying Centralization to Two Detectors}
\label{sec:existing}
Here we show that Subgraph Centralization is applicable to an existing subgraph-based detector CoLA \cite{liu2021anomaly}, and
an existing full-graph-based detector OCGNN \cite{OCGNN}. They are described in the next two subsections.

\subsection{Centralize a Subgraph-based Detector - CoLA.}

\begin{figure}

 \center
   
  \includegraphics[width=\columnwidth]{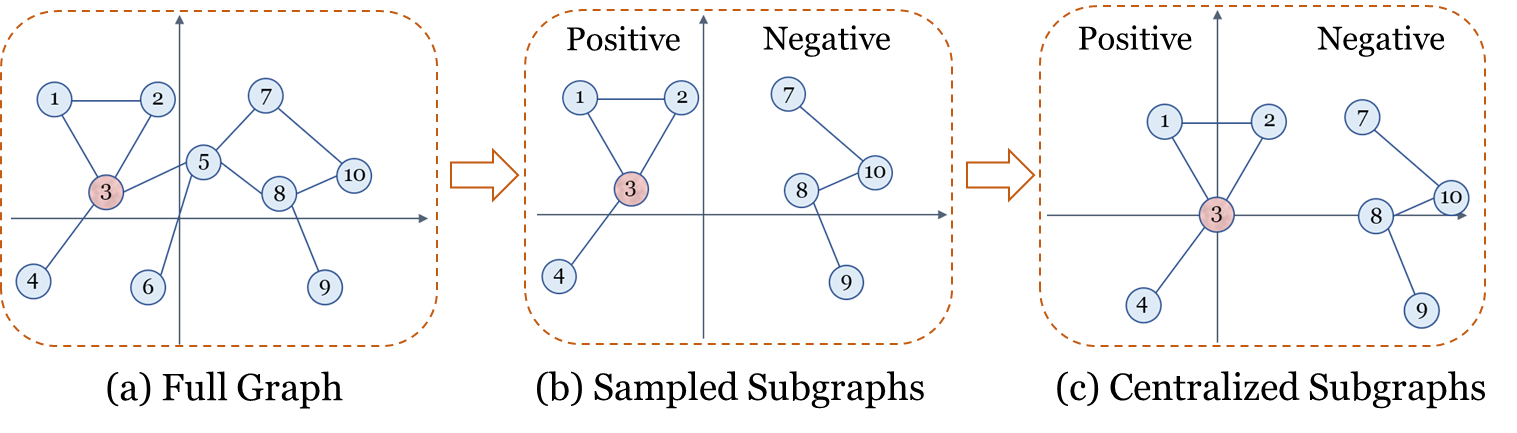}
  \caption{CoLA\_SC performs Subgraph Centralization after the subgraphs have been sampled from a network. The red node denotes the target node in this example.}
   \label{fig:CoLA}
 
\end{figure}

Contrastive self-supervised Learning framework for Anomaly detection on attributed networks (CoLA) \cite{liu2021anomaly} is a recent state-of-the-art graph anomaly detection that uses contrastive learning of subgraphs to learn a classifier so that it can be used to produce an anomaly score for each node. 
We show below that our subgraph centralization can be easily plugged into CoLA to enhance its detection accuracy by performing Subgraph Centralization to CoLA's subgraphs before learning a classifier. Particularly, as shown in Figure~\ref{fig:CoLA}, the new version called CoLA\_SC takes three steps to obtain the centralized subgraphs. The first step is to traverse each node in random order as the target node within every epoch. Then, for a given target node, a positive subgraph and a negative subgraph are sampled via random walk with restart (RWR) \cite{randomwalk}. The positive subgraph is generated from the target node; and the negative subgraph is from a node different from the target node. The first two steps are the same as the way in \cite{liu2021anomaly}. In the third step, both the positive and negative subgraphs are centralized as the way in Algorithm \ref{alg:Subgraph}. After obtaining the centralized subgraphs, exactly the same training steps are used to learn a CoLA detector. The details can be found in \cite{liu2021anomaly}.

\subsection{Centralize a  Full-Graph-based Detector - OCGNN.}
One-class graph neural network (OCGNN) \cite{OCGNN} integrates GNNs and one-class hypersphere learning to provide an end-to-end detector. The enhanced version with Subgraph Centralization is called OCGNN\_SC. The additional functionalities are given as follows. Before training the end-to-end detector, random walk \cite{randomwalk} is used to generate a subgraph for each node, and then each subgraph is centralized to the origin of the node vector space, as described before. An average pooling function is employed as the readout function for GNN, followed by the same hypersphere learning as in OCGNN.

\section{Experimental Design and Results}
\label{sec:experiment}
The experiments are designed to answer the following questions:
\renewcommand{\theenumi}{\roman{enumi}}
\begin{enumerate}
    \item Is Subgraph Centralization a necessary step in the problem of graph anomaly detection?
    \item How does Subgraph Centralization derive its explainability?

\end{enumerate}

To answer the first question, we use the following
baseline methods:
\begin{itemize}
    \item \textbf{Oddball} \cite{10.1007/978-3-642-13672-6_40} is a detector specially designed to detect the types of node anomalies which have clique and star. It learns a power-law model of a network to represent the normal neighbourhoods, and computes the deviation of each node from the expected value of the model as the anomaly score.
    \item \textbf{DOMINANT} \cite{ding2019deep} is a GCN-Autoencoder-based method, denoted as DOM. It applies two decoders to reconstruct the attribute and adjacency matrix. The anomaly score is computed by combining two reconstruction errors with a trade-off coefficient.
    \item \textbf{CoLA} \cite{liu2021anomaly} is a contrastive learning method that trains a GCN-based model to discriminate node anomalies from normal nodes in a network
    \item \textbf{OCGNN} \cite{OCGNN} is a method utilizing GCN \cite{Kipf:2016tc} and hypersphere learning \cite{SVDD} to detect anomalies.
\end{itemize}

In GCAD \footnote{Code is available at \url{https://github.com/IsolationKernel/Codes/tree/main/IDK/GraphAnomalyDetection.}}, we use the recent Isolation Distributional Kernel  \cite{10.1145/3394486.3403062} (IDK for short) as the point anomaly detector in line\#6 in Algorithm~1. We also examine alternative existing unsupervised anomaly detectors in an ablation study later. And other experimental settings are given in the Appendix.

\setlength{\tabcolsep}{2pt}
\begin{table}[h]
\centering
\resizebox{1.0\columnwidth}{!}{%
\begin{tabular}{ccrrrr}
\hline \hline

 \multicolumn{1}{c|}{}                & \# Nodes & \# Edges & \# Attributes & \# Anomalies  \\ \hline
 \multicolumn{1}{c|}{Watts-Strogatz}            & 500     & 1500    & 2            & 24             \\
 \multicolumn{1}{c|}{SBM$_{stru}$}            & 1000    & 5839    & 10           & 25              \\ 
 
 \multicolumn{1}{c|}{RGG$_{s}$}                     & 500     & 2195    & 2            & 20                 \\
\multicolumn{1}{c|}{RGG$_{l}$}                & 500     & 60639  & 2            & 20                   \\
\multicolumn{1}{c|}{Lattice$_l$}                 & 1200    & 2340    & 2            & 20                     \\
\multicolumn{1}{c|}{Lattice$_s$}                  & 1220    & 2410    & 2            & 99                \\ \hline

\multicolumn{1}{c|}{ACM}                 & 16484   & 82175   & 8337         & 597                 \\

\multicolumn{1}{c|}{Cora}                  & 2708    & 5803    & 1433         & 150               \\

\multicolumn{1}{c|}{Citeseer}           & 3327    & 5139    & 3703         & 150               \\

\multicolumn{1}{c|}{Pubmed}              & 19717   & 46424   & 500          & 600                \\ \hline

\hline \hline
\end{tabular}%
}
\caption{Data characteristics of datasets used}
\label{tab:my-table}
\end{table}

\setlength{\tabcolsep}{2pt}
\begin{table}[h]
\centering
\resizebox{1\columnwidth}{!}{%
\begin{tabular}{c|rr|rr|rr|r}
\hline \hline

 & Oddball  & DOM & \multicolumn{2}{c|}{CoLA}   & \multicolumn{2}{c|}{OCGNN} & GCAD   \\ \cline{4-7}
 &   &  & Orig & SC  & Orig & SC &    \\ \hline
 \multicolumn{1}{c|}{Watts-Strogatz}  &  .63    &  .63   &  .72           &     \textbf{.82} & .24  &  .77 & \textbf{.82}      \\
 \multicolumn{1}{c|}{SBM$_{stru}$}            & \textbf{1.00}    &  \textbf{1.00}   &     \textbf{1.00}      &  \textbf{1.00} &  .21 &\textbf{1.00} &\textbf{1.00}     \\ 
 
 \multicolumn{1}{c|}{RGG$_{s}$}                     &  .56    & .70    &   .70          &  .84   & .43 &      .86   & \textbf{.91}    \\
\multicolumn{1}{c|}{RGG$_{l}$}                    & .99     & .93  &\textbf{1.00}          &\textbf{1.00} &.40  & .40      & \textbf{1.00}   \\
\multicolumn{1}{c|}{Lattice$_l$}                 & .94    & .95    & .80            & .83         & .30    &\textbf{1.00}  & \textbf{1.00}   \\
\multicolumn{1}{c|}{Lattice$_s$}             &\textbf{1.00}  & .92    & .65            & .73     &.55        & .92  & .96 \\ \hline
\multicolumn{1}{c|}{ACM}                  & .75   & .81   & .82         & \textbf{.88}          & .37  & .79 & .85    \\

\multicolumn{1}{c|}{Cora}                & .77    & .90    & .88         & .90         & .24    & .88  &\textbf{.92} \\

\multicolumn{1}{c|}{Citeseer}            & .73    & .93    & .90         & .92         & .26   & .93  & \textbf{.95} \\

\multicolumn{1}{c|}{Pubmed}               & .74   & .91   & \textbf{.95}          & \textbf{.95}       & .37   & .89    & .92  \\ \hline
\multicolumn{1}{r|}{Rank}             & 4.8   &       3.85    &   3.85    &  2.95       &   6.95    &       3.7   & 1.85 \\ \hline
\multicolumn{1}{r|}{P-value}& .0076 & .0038 & .0086 & .1020 & .0010 & .0059  &-\\

\hline \hline
\end{tabular}%
}
\caption{Detection accuracy in terms of AUC of different methods. `Orig' denotes the original method; and `SC' denotes a method is modified with Subgraph Centralization.  The last row shows the p-value obtained from a pair-wise significance test in comparison with GCAD. The second last row shows the ranking of each method, average all datasets.}
\label{tab:my-table}
\end{table}

The investigation for the first question is reported in Section \ref{sec:key-results}. The answer to the second question is presented in Section \ref{sec:explainability}. Two additional experiments, that examine the effect of $h$ in $h$-subgraphs and a scaleup test, are reported in the following two subsections.

\subsection{Main Evaluation.}
\label{sec:key-results}

\begin{table}[h]
\centering
\tiny
\resizebox{1.0\columnwidth}{!}{%
\begin{tabular}{c|cc}
\hline
 & Normal & Abnormal  \\ \hline
\multicolumn{1}{c|}{Watts} & \begin{minipage}[b]{0.25\columnwidth}
     \centering
     \raisebox{-0.5\height}{\includegraphics[width=\linewidth]{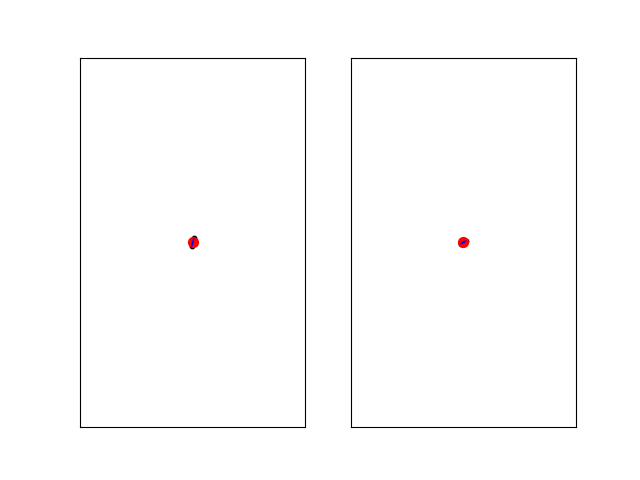}}
    \end{minipage} & \begin{minipage}[b]{0.25\columnwidth}
     \centering
     \raisebox{-0.5\height}{\includegraphics[width=\linewidth]{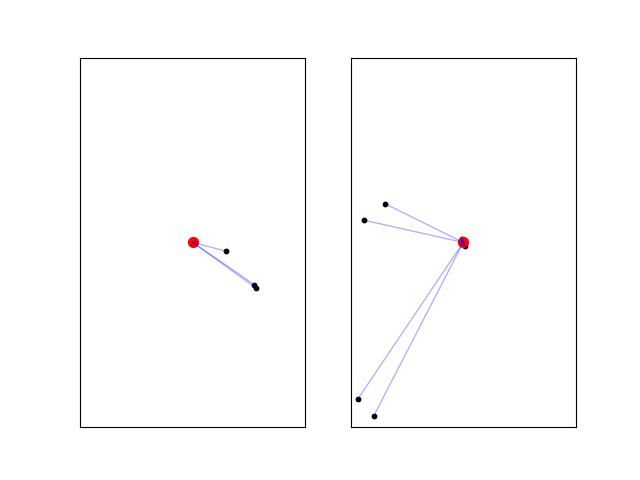}}
    \end{minipage} 
   
    \\ \hline
\multicolumn{1}{c|}{RGG$_s$} & \begin{minipage}[b]{0.25\columnwidth}
     \centering
     \raisebox{-0.5\height}{\includegraphics[width=\linewidth]{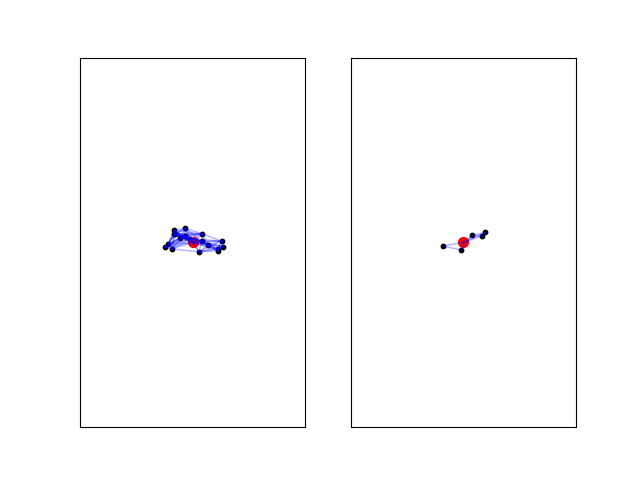}}
    \end{minipage} & \begin{minipage}[b]{0.25\columnwidth}
     \centering
     \raisebox{-0.5\height}{\includegraphics[width=\linewidth]{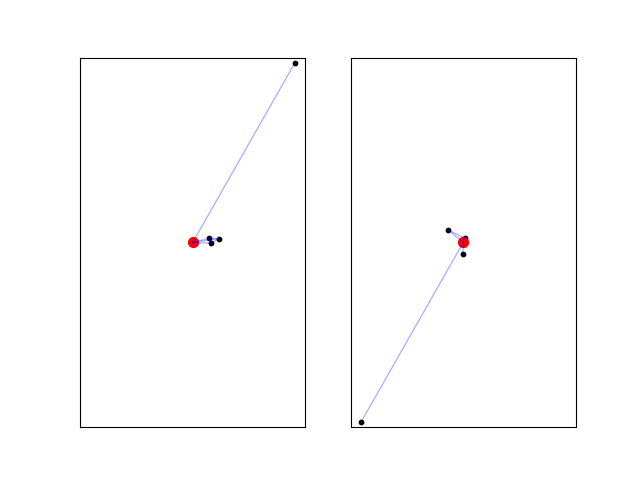}}
    \end{minipage}
    
    \\ \hline
\multicolumn{1}{c|}{RGG$_l$} & \begin{minipage}[b]{0.25\columnwidth}
     \centering
     \raisebox{-0.5\height}{\includegraphics[width=\linewidth]{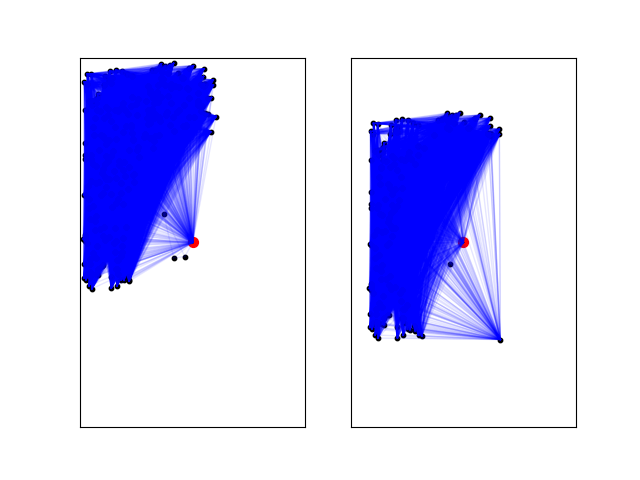}}
    \end{minipage} & \begin{minipage}[b]{0.25\columnwidth}
     \centering
     \raisebox{-0.5\height}{\includegraphics[width=\linewidth]{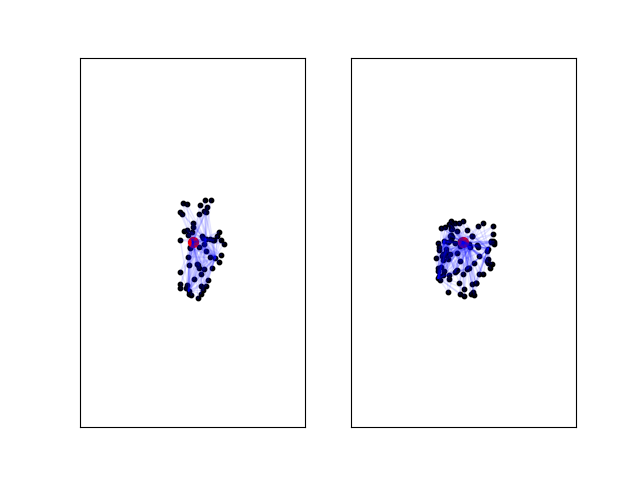}}
    \end{minipage}
    \\ \hline
\multicolumn{1}{c|}{Cora} & \begin{minipage}[b]{0.25\columnwidth}
     \centering
     \raisebox{-0.5\height}{\includegraphics[width=\linewidth]{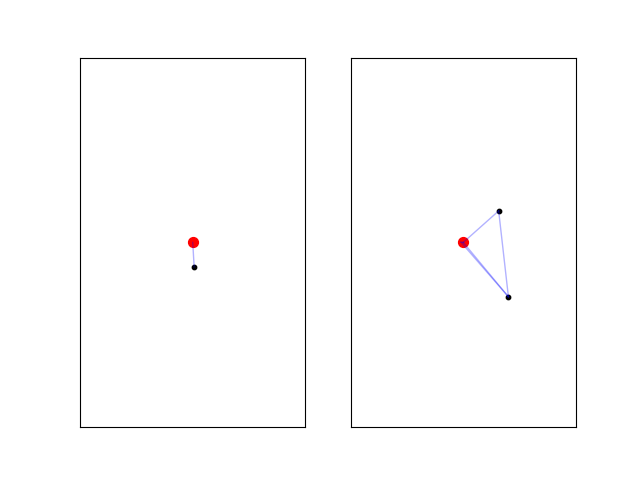}}
    \end{minipage} & \begin{minipage}[b]{0.25\columnwidth}
     \centering
     \raisebox{-0.5\height}{\includegraphics[width=\linewidth]{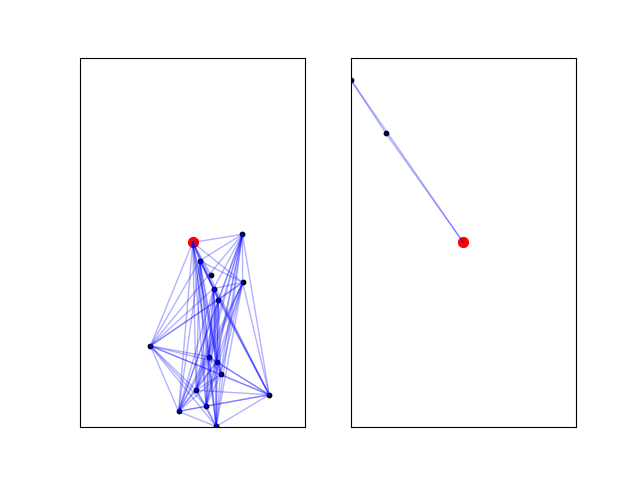}}
    \end{minipage}
    \\ \hline
\end{tabular}%
}
\caption{Examples of anomalous and normal nodes identified by GCAD via $h$-subgraphs. Each red node denotes the source node of a centralized $h$-subgraph. These visualizations uses $1$-subgraphs only. Note that many (normal) nodes of RGG$_l$ have many long connections and the few anomalous ones have short connections, unlike those in the other three datasets.}
\label{tab:explaination}
\end{table}

This evaluation employs two sets of datasets, i.e., synthetic datasets and existing datasets. Their data characteristics are given in Table \ref{tab:my-table}. They are briefly described as follows.

\textbf{Synthetic datasets}. We create six synthetic datasets using four classical graph generation models \cite{2011,watts,2002,conway1998sphere}. For RGG and Lattice, we create two types of datasets, whose normal node in one is the anomalous node in the other. Such setting can help us evaluate the ability of the methods in detecting different types of anomalies apart from the type found in the real datasets. (See the Appendix for details).

\textbf{Existing datasets}. 
Like previous work, the original citation network datasets, i.e., Cora, Citeseer, Pubmed and ACM, are assumed to have no anomalies and two types of anomalies are injected. They are the same as used in \cite{liu2021anomaly}.\\

\textbf{Overall results}. Observations from  Table \ref{tab:my-table} are:
\begin{itemize}
   
    \item GCAD is the best. It performs the best in seven out of the ten datasets, and the second best in the other three exceptions. It is also significantly better than each of the other six contenders at the $p=0.01$ significance level. The only exception is CoLA\_SC which employs the Centralization. 
    \item Subgraph Centralization significantly improves the performance of CoLA and OCGNN close to the level of GCAD. The improvement over OCGNN is huge on almost all datasets\footnote{There is no improvement for OCGNN$\_$SC on RGG$_l$ owing to the fundamental limitation of SVDD, i.e., SVDD always regards points far away from the data centroid as anomalies. This dataset has anomalies at the data centroid and the normal points are further away from the data centroid. \label{fn-SVDD}}. Similarly, the improvement over CoLA is on all datasets that have room for improvement, albeit the degree of improvement is smaller. 
\end{itemize}

\subsection{Explainability.}
\label{sec:explainability}
Table \ref{tab:explaination} shows examples of anomalous and normal nodes identified by GCAD via $h$-subgraphs. This explanation via visualization is possible because of the use of Subgraph Centralization before the anomaly detection. The top two normal $h$-subgraphs indicate the typical $h$-subgraphs that exist in a network (ranked at the bottom in the ranked list). The top two anomalous $h$-subgraphs indicate the most unusual $h$-subgraphs in the network. The examples in Table \ref{tab:explaination} demonstrate that the $h$-subgraphs of the detected anomalies are significantly different from those normal $h$-subgraphs in each dataset. 

Interestingly,  \emph{this explainability depends on the use of centralized subgraphs only}, and is independent of the point anomaly detector used in line\#6 in Algorithm \ref{alg:framework}. 

As none of the existing three detectors, i.e., CoLA, DOMINANT and OCGNN employ Subgraph Centralization for detection, they are unable to provide such a visualization to explain their detection outcomes. However, after utilizing Subgraph Centralization, CoLA\_SC and OCGNN\_SC have the same explainability.

\subsection{The Effect of Parameter $h$ on GCAD}
\label{sec:h}
is examined here. GCAD performs the best when $h=1$ on nine out of ten datasets and the only exception is RGG$_s$. This means that the simplest 1-subgraphs are sufficient to detect the anomalies on almost all the datasets. Figure \ref{fig:h} in the Appendix shows the effect of $h$ on different datasets.

\subsection{Scaleup Test and Time Complexity.}

\begin{figure}[h]
\vspace{-5mm}
 \center

  \includegraphics[width=1.0\linewidth]{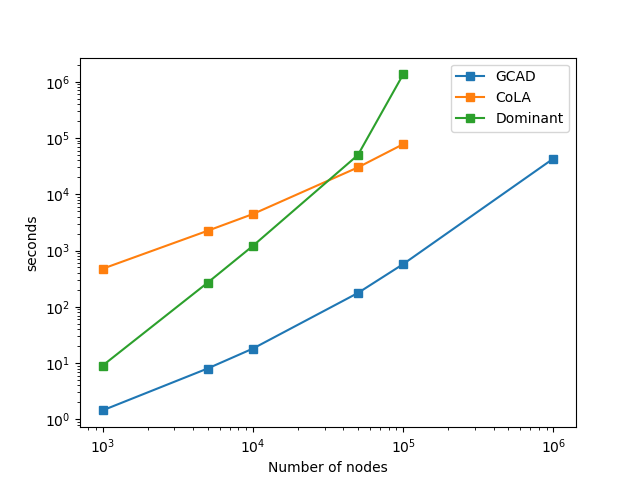}
\vspace{-3mm}
  \caption{Scaleup test for GCAD, CoLA, and DOMINANT on the Watts-Strogatz dataset.}

  \label{scaleup}

\end{figure}

Figure \ref{scaleup} shows the result of a scaleup test for GCAD, DOMINANT and CoLA on Watts-Strogatz where the number of nodes is increased from $10^3$ to $10^6$. For a fair comparison, we measure the time each algorithm takes under the same environment. \\
The result shows that GCAD needs far fewer computations and can deal with large-scale million-node networks. DOMINANT and CoLA took a prohibitively long time when applied to the million-node network.\\
The time complexities of different components of GCAD are given in Table \ref{tab:Time-complexity}. GCAD has linear time complexity since $m \ll n$ and the other parameters are constant.

Note that the time complexities for CoLA and OCGNN  do not include the deep learning time.

\begin{table}[h]
    \centering
    \resizebox{0.95\columnwidth}{!}{%
    \begin{tabular}{c|l|c}
    \hline
\multirow{5}{*}{GCAD} & Subgraph Extraction &  $\mathcal{O}(nm)$\\
 &Subgraph Centralization & $\mathcal{O}(nmf)$\\
&Subgraph Embedding (WL scheme \cite{Togninalli19}) & $\mathcal{O}(feh)$\\
&Detector IDK \cite{10.1145/3394486.3403062} & $\mathcal{O}(nt\psi)$\\ \cline{2-3}
&Total of GCAD & $\mathcal{O}(n(mf + t\psi) + feh)$\\ \hline
\multirow{1}{*}{CoLA} & Subgraph Generation & $\mathcal{O}(mnR(m + \delta))$\\ \hline
\multirow{1}{*}{OCGNN} & One GCN layer  & $\mathcal{O}(pfu)$\\ \hline
    \end{tabular}
    }
    \caption{The time complexities of different components in GCAD. $f$ and $n$ denote the number of dimensions (of node vectors) and the number of nodes of a network. $m$ and $e$ denote the maximum numbers of nodes and edges, respectively, in a subgraph. $R$ and $\delta$ denote the sampling rounds and the average degree in a network, respectively. $p$ and $u$ are the number of non-zero elements in adjacency matrix and the number of feature maps of the weight matrix, respectively.}
    \label{tab:Time-complexity}
\end{table}

\section{Ablation Studies on GCAD}
\label{sec:ablation}

\begin{table}[]
\centering

\resizebox{1.0\columnwidth}{!}{%
\begin{tabular}{c|rrrr|rrr}
\hline\hline
 & \multicolumn{4}{c|}{IDK}  & iForest & OCSVM & LOF   \\\hline 
Centralization   &   \XSolidBrush    &  \Checkmark     &    \XSolidBrush   &    \Checkmark   &    \Checkmark     &     \Checkmark  &     \Checkmark  \\
Weighted score   &    \XSolidBrush   &    \XSolidBrush   &     \Checkmark  &  \Checkmark     &    \Checkmark     &  \Checkmark     &    \Checkmark   \\ \hline

Watts-Strogatz   & .57 & \textbf{.82} & .66 & \textbf{.82} & \textbf{.82}   & \textbf{.82} & \textbf{.82} \\
SBM$_{stru}$     & .58 & \textbf{1.00}   & .60 & \textbf{1.00}   & \textbf{1.00}     & .99 & \textbf{1.00}  \\
RGG$_s$          & .56 & .90 & .59 & .91 & .92   & .93 & \textbf{.95} \\
RGG$_l$          & .88 & \textbf{1.00}   & .89 & \textbf{1.00}  & .51   & .52 & \textbf{1.00} \\
Lattice$_l$      & .57 & \textbf{1.00}   & .60 & \textbf{1.00}   & \textbf{1.00}     & \textbf{1.00}   & .98 \\
Lattice$_s$      & .50 & .96 & .51 & .96 & \textbf{.98}   & .96 & .97 \\
 \hline
ACM              & .73 & .82 & .76 & \textbf{.85} & .71   & \textbf{.85} & .83 \\
Cora             & .64 & .90 & .64 & \textbf{.92} & .85   & \textbf{.92} & \textbf{.92} \\
Citeseer         & .59 & .94 & .59 & \textbf{.95} & .81   & \textbf{.95} & \textbf{.95} \\
Pubmed           & .69 & .89 & .74 & .92 & .82   & \textbf{.94} & .87 \\ \hline

Rank             &    6.6   &  3.3     & 5.8       &    2.5   &     4.1    &    2.85   &     2.85  \\ \hline
P-value          &  .0010     &  .0206     &   .0010    &    -    &      .0315    &    .5     &   .1238\\ \hline \hline
\end{tabular}%
}
\caption{Results of two ablation studies on GCAD w.r.t. subgraph centralization, weighted score and point anomaly detectors in terms of AUC. }
\label{tab:ablation}
\end{table}
In this section, we conduct two ablation studies. The first examines the effects of the two components in GCAD, i.e., Subgraph centralization and Depth-based weighted score. The second investigates the generic nature of the GCAD framework by using different existing point anomaly detectors in line\#6 in Algorithm 1. The unsupervised point anomaly detectors investigated are Isolation Forest (iForest) \cite{4781136}, OCSVM \cite{ocsvm}, and Local Outlier Factor (LOF) \cite{LOF} and IDK \cite{10.1145/3394486.3403062} (we have used IDK as the default in the previous sections).  

The results of the first study are shown in the second to the fifth columns in Table \ref{tab:ablation}; and the results of the second study are shown in the fifth to the last columns. 

We summarize the outcomes of these studies as: 
\begin{itemize}
    \item \textbf{Subgraph centralization} is a crucial component of GCAD. The accuracies of GCAD decline significantly on all the datasets without the component.
    \item \textbf{Weighted score}. The weighted score component provides a small AUC improvement on some datasets.
 It is useful especially when anomalous nodes are connected such as cliques. Through the depth-based weighted score, an anomalous node increases the anomaly scores of its anomalous neighbours. Therefore, all the anomalous nodes are ranked higher than without the weighted score.
    \item \textbf{Point anomaly detectors}. The results in the last three columns in Table \ref{tab:ablation} show that IDK performs better than iForest at $p=0.05$ significance level. Overall, IDK has comparable accuracy as OCSVM and LOF. OCSVM performs poorly on RGG$_{l}$ for the same reason  stated in footnote \ref{fn-SVDD} wrt OCGNN in Section~\ref{sec:key-results} because they make the same assumption. The example runtimes on the Pubmed dataset of GCADs with IDK, iForest, OCSVM and LOF are  23, 40, 588 and 583 seconds, respectively. IDK is the recommended point anomaly detector at this point in time because it has the highest accuracy and run fastest. 
\end{itemize}

\section{Discussion}
\label{sec:discussion}

\textbf{Anomalous Node Definitions}.
The importance of the definition of anomalous nodes can not be under-estimated.  Without a clear definition, how an anomaly  relates to the normal ones is unclear. We cannot find such a definition in all the current works listed in the related work section. This has two implications. First, the analysis of the reason why (or the condition under which) a detector work or not becomes difficult, if not impossible. Second, anomalies are often implicitly/explicitly assumed to belong to a very restricted kind in existing works. For example, a node anomaly is assumed to be far away from an existing node in a network \cite{ding2019deep, liu2022bond}. This assumption is made on the basis that normal nodes are close to each other. 

One crucial ingredient is missing in the above assumption, i.e., anomalies are rare and normal ones are plentiful in a network. It is possible that a network may contain two (or more) types of normal nodes in different parts. For example, one part has many subgraphs with close neighbouring nodes, and the other has many subgraphs with far away neighbouring nodes. As a result, anomalies in one part becomes normal ones in the other. Using the assumption misses all the anomalies in one part of the network, as we have identified with OCGNN in Section~\ref{sec:key-results}. 

The use of Definition \ref{def:normal} for anomalous/normal nodes assumes that a similarity measure is used in GCAD's detector. IDK is a measure as well as a detector.  Our empirical result in Table \ref{tab:ablation} shows that using IDK has an advantage over other detectors in terms of both anomaly detection accuracy and computational time.

Note that the GCAD framework is generic, not coupled with any particular detector (as reported in Section \ref{sec:ablation}). It is also not tied to any specific similarity measure, and it even admits no measure.

In fact, the use of a different point anomaly detector in the GCAD framework changes the definition of node anomalies. This is analogous to the definition of point anomalies in point anomaly detection.  Definition \ref{def:normal} is valid when GCAD employs IDK. When GCAD uses iForest, the definition shall be revised to: Node $u$ in $\mathcal{G}$ is an anomalous node if $\phi(\mathcal{G}^{h}(u))$ is susceptible to isolation.

\noindent
\textbf{Anomaly Types in a Graph}.
We have identified that the  commonly used datasets are likely to belong to one type of anomaly, where only relative positions of node vectors matter. That is why the proposed method works better than existing methods which do not employ subgraph centralization. It is possible that there are other types of graphs in which the proposed method does not work. Two possible example types are that anomalous nodes are due to (a) node vectors only but not graph structure, and (b) the absolute positions of node vectors play an important role in addition to graph structure. The former is not a graph problem and the node anomalies can be identified with a point anomaly detector. The latter is an open problem.

\section{Conclusion}
\label{sec:conclusion}
We show that Subgraph Centralization is a necessary technique in graph anomaly detection that has a significant influence on detection accuracy, and it empowers any detectors to gain explainability that would otherwise be impossible. As this technique has time complexity linear to the number of nodes, it enables a detector to deal with large scale datasets.
These advantages are shown in our proposed framework GCAD that incorporates Subgraph Centralization. Our empirical evaluation verifies that (i) GCAD is superior to four existing detectors in terms of its detecting accuracy, run time and scalability; and (ii) two existing detectors CoLA and OCGNN can reap at least two out of the three above-mentioned benefits of Subgraph Centralization if it is integrated into their algorithms.

\section*{Acknowledgement}
Kai Ming Ting is supported by National Natural Science Foundation of China (62076120). Guansong Pang is supported in part by the Singapore Ministry of Education
(MoE) Academic Research Fund (AcRF) Tier 1 grant (21SISSMU031).

\bibliography{sample-base.bib}
\bibliographystyle{plain}

\newpage

 \section*{Appendix}

This Appendix provides the experimental settings used in this paper, and some additional results.

\subsection*{Experimental Settings:}
We use the pairwise non-parametric Wilcoxon Signed-Rank Test \cite{10.2307/3001968} to quantify the differences between the proposed GCAD and each of the four baselines.
The detection accuracy of an anomaly detector is measured in terms of AUC (Area under ROC curve) which is widely used in most papers on anomaly detection.\\

\subsection*{Synthetic Datasets:}
\label{Synthetic}
We create several datasets using classical graph generation models including Block Model \cite{2011}, Watts–Strogatz Model \cite{watts}, Random Geometric Graph \cite{2002} and Lattice \cite{conway1998sphere}, and label nodes with abnormal properties as anomalies as:
\begin{itemize}
    \item \textbf{Stochastic Blockmodels} \cite{2011}. We create $n$ nodes, and a node is assigned to one of the $t$ blocks in the network. The probability that an edge exists between a node of a certain group to another node of the same group (or a different group) is fixed and independent. For each block, the node vectors are distributed following a Gaussian distribution. In SBM$_{stru}$, several cliques are injected into some randomly chosen nodes.
    \item \textbf{Random Geometric Graphs} \cite{2002}. $d$-dimensional random geometric graph (RGG) is a graph where each of the $n$ nodes is assigned random node vectors with a uniform distribution in a $d$-dimensional box. In RGG$_s$, given a threshold $\tau$, normal nodes connect nodes whose Euclidean distance is smaller than $\tau$, while the anomalous nodes connect those whose Euclidean distance is larget than $\tau$. RGG$_l$ is contrary to RGG$_s$. We create these two graphs to show that it is not always the case that nodes with distant connections are anomalies;
    
    \item \textbf{Lattice} \cite{conway1998sphere}. This is a squared lattice in a two-dimensional space. In Lattice$_l$, an anomaly is generated via connecting pairs of unconnected nodes. In Lattice$_s$, an anomaly is injected by randomly generating node vectors in the node vector space (which fall within the lattice squares) and connect the injected nodes with the four nearest nodes.
    \item \textbf{Watts-Strogatz} \cite{watts}. The Watts-Strogatz generator generates random small-world graphs. We generate a graph which has an average node degree of 6 and 500 nodes. The nodes which have node degree $\geq 8$ or $\leq 4$ are regarded as anomalies. 
\end{itemize}

\subsection*{Parameter Search Range:}
The search ranges of the parameters in the experiments are shown in Table \ref{parameter}.

\begin{table}[h]
\centering
\resizebox{0.7\columnwidth}{!}{%
    \begin{tabular}{@{}ll@{}}
    \hline
        Algorithm & Parameter Search Range \\ \hline
        GCAD & $h\in\{1,2\}$; $\psi_{IDK}\in\{2,4,8\}$;\\ & $\lambda\in\{2^{-1}, 2^{-2},\dots, 2^{-5}\}$\\
        DOMINANT & $\alpha \in \{0.0, 0.1,\dots, 1.0\}$\\
        CoLA & $R\in\{64, 128, 256\}$; $c\in\{2,3,\dots,8\}$; \\ & $d\in\{2,4,\dots,64\}$\\
        OCGNN & $\lambda\in\{0.05, 0.01, 0.0005\}$; hidden size$=64$;\\
        \hline
    \end{tabular}
    }
    \caption{Parameter search ranges in the experiments}
    \label{parameter}
\end{table}

\subsection*{Loss Curve:}
\begin{figure}[h]

 \center

  \includegraphics[width=.7\linewidth]{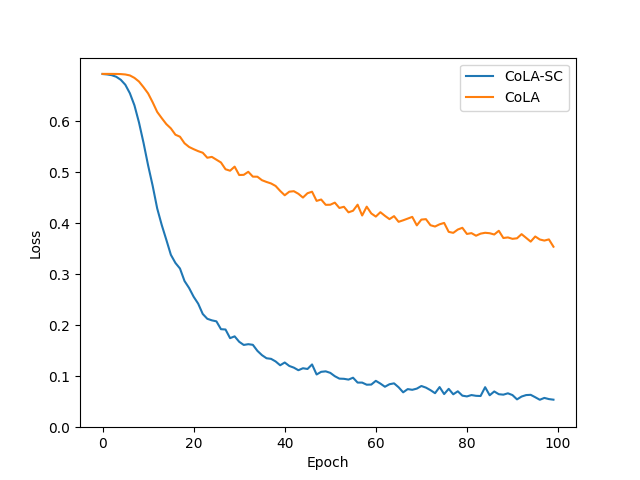}
\vspace{-3mm}
  \caption{Loss curves of CoLA and CoLA\_SC}

  \label{fig-loss}

\end{figure}

Figure \ref{fig-loss} shows the progression of loss values as the learning epochs continue for both CoLA and CoLA\_SC. This result shows that the impact of centralized subgraphs enables the learning to converge much faster than using the uncentralized subgraphs.

\subsection*{Effect of $h$ on GCAD:} 
\begin{figure}[h]

  \centering
  \includegraphics[width=.7\linewidth]{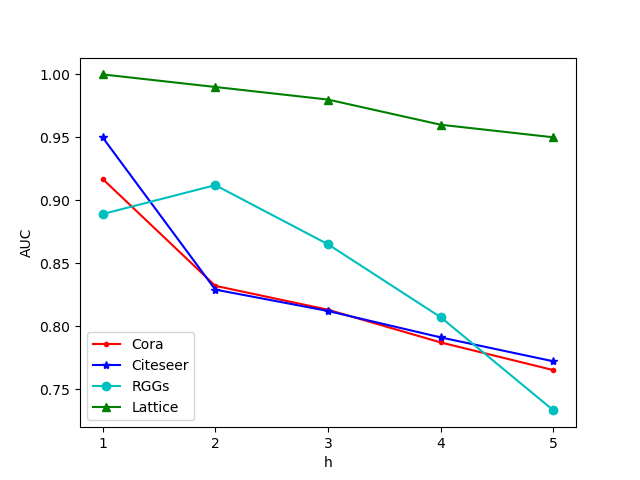}
\vspace{-3mm}
  \caption{The effect of $h$ on GCAD on different datasets.}

  \label{fig:h}

\end{figure}

As shown in Figure \ref{fig:h}, $h=1$ is required on most datasets, while a larger $h$ is needed on some other datasets.

\end{document}